\documentclass[letterpaper]{article} 
\usepackage{aaai24}  
\usepackage{times}  
\usepackage{helvet}  
\usepackage{courier}  
\usepackage[hyphens]{url}  
\usepackage{graphicx} 
\usepackage{natbib}  
\usepackage{caption} 
\usepackage{algorithm}
\usepackage{algorithmic}
\usepackage{array}
\usepackage{amsmath}
\usepackage{amssymb}
\usepackage{booktabs} 
\usepackage{multirow}
\usepackage{bbding}
\usepackage{subcaption}
\usepackage{appendix}
\newcolumntype{L}{>{\fontsize{9}{10}\selectfont}l}
\newcommand{\sdfnetwork}{F_g}
\newcommand{\colornetwork}{F_c}
\newcommand{\seghead}{F_s}
\newcommand{\view}{d}
\newcommand{\surface}{\mathcal{S}}
\newcommand{\pixel}{p}
\newcommand{\origin}{o}
\newcommand{\nframe}{k}
\newcommand{\shapeparameter}{\beta}
\newcommand{\poseparameter}{\theta}
\newcommand{\hand}{H}
\newcommand{\amodal}{O}
\newcommand{\loss}{\mathcal{L}}
\newcommand{\joint}{J}
\newcommand{\vertices}{v}
\newcommand{\image}{I}
\newcommand{\surfacepoint}{x_s}
\newcommand{\surfacenormal}{n_s}
\newcommand{\surfacenormalneig}{n_{s+\epsilon}}
\newcommand{\handsurfacepoint}{v_s}
\newcommand{\distance}{r}
\newcommand{\yuzhi}{\tau}
\newcommand{\tanhxishu}{\beta}
\newcommand{\logit}{l}
\newcommand{\renderlogit}{L}
\newcommand{\seg}{M}
\newcommand{\xishu}{\lambda}
\newcommand{\raynum}{N_r}
\newcommand{\pointnum}{N_p}
\newcommand{\pixelcolor}{C}
\newcommand{\weight}{W}
\newcommand{\segnetwork}{N_{seg}}
\newcommand{\amodalnetwork}{N_{amodal}^{hg}}
\newcommand{\reflabel}{dummy} 
\newcommand{\figlabel}[2][\reflabel]{\label{fig:#1-#2}}
\newcommand{\figref}[2][\reflabel]{Fig.{~\ref{fig:#1-#2}}}
\newcommand{\seclabel}[1]{\label{sec:\reflabel-#1}}

\newcommand{\eqlabel}[1]{\label{eq:\reflabel-#1}}
\newcommand{\Eqref}[2][\reflabel]{Eq.~{\ref{eq:#1-#2}}}
\newcommand{\tablelabel}[2][\reflabel]{\label{table:#1-#2}}
\newcommand{\tableref}[2][\reflabel]{Table~{\ref{table:#1-#2}}}
\usepackage{newfloat}
\usepackage{listings}
\DeclareCaptionStyle{ruled}{labelfont=normalfont,labelsep=colon,strut=off} 
\floatstyle{ruled}
\newfloat{listing}{tb}{lst}{}
\floatname{listing}{Listing}
\title{In-Hand 3D Object Reconstruction from a Monocular RGB Video}
\author{
    Shijian Jiang\textsuperscript{\rm 1}, Qi Ye\textsuperscript{\rm 1,\rm 2}\thanks{Corresponding author.}, Rengan Xie\textsuperscript{\rm 3}, Yuchi Huo\textsuperscript{\rm 3,\rm 4},
    Xiang Li\textsuperscript{\rm 5}, Yang Zhou\textsuperscript{\rm 5}, Jiming Chen\textsuperscript{\rm 1}
}
\affiliations{
    \textsuperscript{\rm 1}College of Control Science and Engineering, Zhejiang University\\
    \textsuperscript{\rm 2}Key Lab of CS\&AUS of Zhejiang Province\\
    \textsuperscript{\rm 3}State Key Lab of CAD\&CG, Zhejiang University\\
    \textsuperscript{\rm 4}Zhejiang Lab\\
    \textsuperscript{\rm 5}OPPO US Research Center\\
    \{jsj630, qi.ye, rgxie\}@zju.edu.cn, huo.yuchi.sc@gmail.com,\\
    \{xiang.li, yang.zhou\}@oppo.com, cjm@zju.edu.cn
}

\begin{document}

\maketitle
\begin{abstract}
Our work aims to reconstruct a 3D object that is held and rotated by a hand in front of a static RGB camera. Previous methods that use implicit neural representations to recover the geometry of a generic hand-held object from multi-view images achieved compelling results in the visible part of the object. However, these methods falter in accurately capturing the shape within the hand-object contact region due to occlusion. In this paper, we propose a novel method that deals with surface reconstruction under occlusion by incorporating priors of 2D occlusion elucidation and physical contact constraints. For the former, we introduce an object amodal completion network to infer the 2D complete mask of objects under occlusion. To ensure the accuracy and view consistency of the predicted 2D amodal masks, we devise a joint optimization method for both amodal mask refinement and 3D reconstruction. For the latter, we impose penetration and attraction constraints on the local geometry in contact regions. We evaluate our approach on HO3D and HOD datasets and demonstrate that it outperforms the state-of-the-art methods in terms of reconstruction surface quality, with an improvement of $52\%$ on HO3D and $20\%$ on HOD. Project webpage: https://east-j.github.io/ihor.
\end{abstract}

\section{Introduction}
3D object reconstruction from images has many applications in fields such as robotic manipulation and AR/VR. A handy and low-cost way to obtain 3D models is to rotate an object in hand in front of a camera and reconstruct the 3D objects from a captured video, which is the focus of this work. However, in-hand 3D object reconstruction in this setting poses several challenges, such as the lack of prior knowledge of the object shape, the estimation of the relative poses between the camera and the object, and particularly, the occlusion caused by the hand-object interaction.

Implicit neural representations, combined with volume rendering techniques~\cite{neus, volsdf}, have proven to be remarkably effective in reconstructing 3D geometry from multi-view images without requiring any prior knowledge of the object. Several in-hand object reconstruction works based on these representations~\cite {hampali2022hand, hhor,wen2023bundlesdf} have achieved compelling results in the visible part of the object. However, their performance degrades significantly when objects are heavily occluded by the hand as these methods optimize 3D object models to fit the observed images only.

In this paper, we argue that dealing with object surface reconstruction under occlusion demands the incorporation of additional priors beyond direct observation. Humans are capable of intuitively elucidating objects under occlusion. Some works \cite{back2022unseen, zhan2020self, zhou2021human} therefore explore large-scale data to learn the capability of occlusion elucidation for images with amodal mask completion. However, leveraging this 2D elucidation capability for multi-view 3D reconstruction is challenging as the amodal masks may be inaccurate and inconsistent across multiple views, especially in the heavily occluded areas. To address this issue, we add a semantic amodal mask head to the implicit 3D reconstruction neural network and refine the masks by jointly optimizing the parameters of both networks.

Though the completed amodal mask can help to constrain a rough global shape of an object, it may not reconstruct local surfaces well, as small changes in the 3D local surfaces might not render apparent changes in the 2D masks. On the other hand, we humans can feel the object shapes and manipulate objects by hands without seeing them and an attempt has been made in \cite{yin2023rotating} that robotic hands can accomplish similar tasks with only simple tactile information (touch object surfaces or not) collected by tactile sensors attached on robotic hands. With this inspiration, we propose to infer the occluded local object surfaces by reasoning about the physical contact between objects and hands: the reconstructed hands and objects must not intersect with each other and must be in contact to enforce friction so that objects will not fall due to gravity. To this end, we introduce penetration and attraction penalties to guide the inference of the occluded surface in contact areas with hands. 

By incorporating the 2D occlusion elucidation and the physical contact priors, we propose a novel in-hand 3D object reconstruction method based on implicit representations from a monocular RGB video sequence. We evaluate our method on two datasets HO3D~\cite{ho3d} and HOD~\cite{hhor}. The experiments show that our method can accurately reconstruct objects in both visible and invisible parts and significantly outperforms the state-of-the-art methods in terms of reconstruction quality. Our contributions can be summarized as follows:
\begin{itemize}
    \item We propose a novel method for implicit hand-held object reconstruction that first leverages priors of 2D occlusion elucidation and physical contact constraints.
    \item  For the 2D occlusion elucidation prior, we introduce an amodal mask head and a joint optimization method for both amodal mask refinement and 3D object reconstruction to ensure the accuracy and view consistency of the predicted amodal masks.
    \item For the physical contact prior, we devise penetration loss and attraction loss to regularize the occluded object surface.
    \item We conduct extensive experiments on HO3D and HOD datasets and demonstrate that our approach outperforms state-of-the-art methods in terms of surface quality, with an improvement of $52\%$ on HO3D and $20\%$ on HOD.
\end{itemize}

\begin{figure*}
  \centering
  \includegraphics[width=0.95\textwidth]{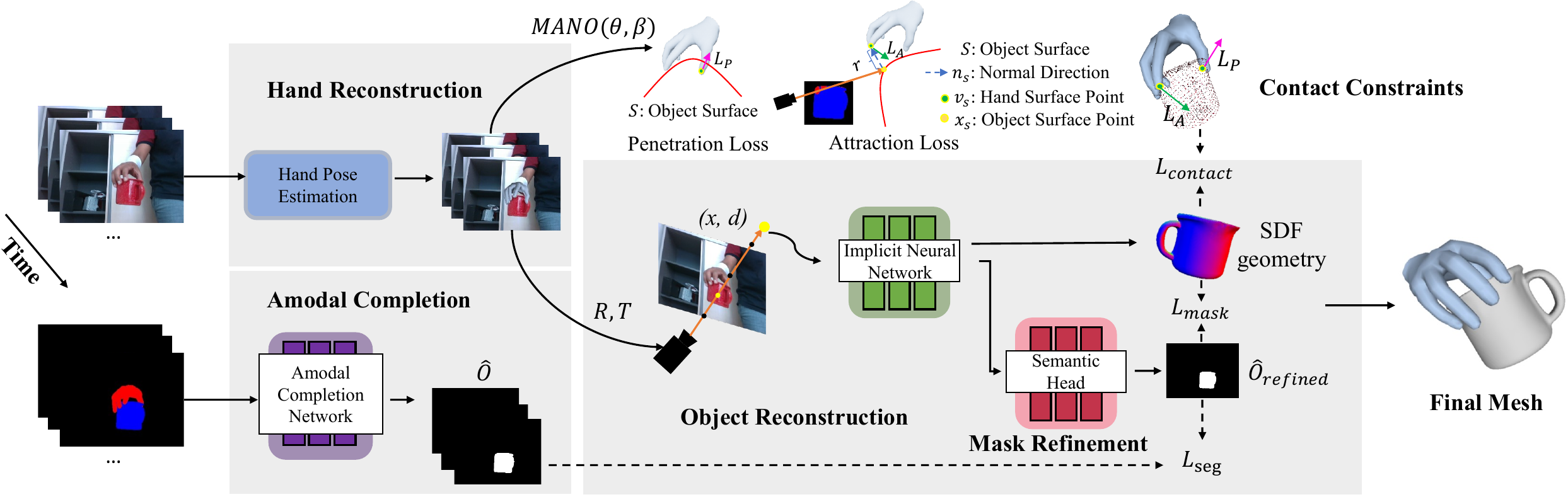}
  \caption{Overview of our framework, consisting of three parts. Hand reconstruction: we optimize the MANO parameters to reconstruct both the hand mesh and camera-relative motion. Amodal completion: by utilizing the amodal completion network, we can recover amodal masks from the input hand and object segmentation maps. Object reconstruction: we learn the 3D objects with the neural implicit field, which is supervised by input images, hand meshes, and amodal masks. To improve the consistency and quality of the predicted amodal masks, we refine them with a semantic head.}
  \figlabel{fig_pipeline}
\end{figure*}

\section{Related Works}
\paragraph{\textbf{Multi-view 3D Reconstruction.}} Recovering 3D geometry from multi-view images has a history in computer graphics and computer vision. Traditional methods~\cite{colmap, barnes2009patchmatch} involve SFM~\cite{sfm} for camera estimation, dense point clouds via MVS, and Poisson reconstruction~\cite{poisson} for meshing. Recently, there is a growing trend to use MLPs to represent 3D appearance and geometry. For instance, NeRF~\cite{nerf} combines the volume rendering with implicit functions by minimizing observed-rendered differences. Inspired by NeRF~\cite{nerf} and SDF~\cite{sdf}, NeuS~\cite{neus} and VolSDF~\cite{volsdf} advance surface quality by replacing the density field with the signed distance fields. We experimentally find these methods tend to produce poor results in the invisible part due to insufficient observation information. Our method is specially designed to handle this occlusion in hand-object interaction scenes.

\paragraph{\textbf{3D Hand-held Objects Reconstruction.}} The 3D reconstruction of manually manipulated objects is a very challenging task due to the heavy occlusion and the variety of objects. To simplify the reconstruction task, several methods~\cite{cao2021reconstructing, yang2022artiboost} reduce the reconstruction to a 6DoF pose estimation. Some other works rely on additional depth information~\cite{zhang2019interactionfusion} or point cloud~\cite{chen2022tracking} to address this challenge. Recent learning-based approaches attempt to directly infer the representations of hands and objects from a monocular RGB image. ~\citet{obman} utilizes AtlasNet~\cite{atlasnet} to recover the object meshes, limited to reconstructing simple objects. ~\cite{ye2022s, graspingfield, alignsdf} use the implicit function to predict the object shape. However, these learning-based methods rely heavily on the dataset, and the reconstructed meshes lack details. In contrast, our method only needs RGB images as supervision and does not need any prior knowledge of the objects. Furthermore, our method excels in recovering object meshes with more details. Most related to our work, ~\cite{hhor, hampali2022hand} reconstructs a hand-held object from RGB monocular video, leveraging the differentiable SDF rendering technique. ~\citet{hhor} treats the interacting hand and object as a whole and separates them using an estimated semantic class of each vertex. ~\citet{hampali2022hand} focus only on the object part. Therefore, these methods do not take occlusion into account, resulting in incomplete surfaces in the hand-occluded part of the object. In contrast, we incorporate contact physical constraints and 2D amodal priors, leading to substantial improvements in the quality of object reconstruction.

\paragraph{\textbf{Occlusion Handling.}} As hands/humans are often severely occluded by objects, several approaches aim to recover the content of the occluded parts. The first approach involves utilizing temporal information. ~\citet{cheng2019occlusion} feed filtered reliable 2D keypoints to 2D and 3D temporal convolutional networks that enforce temporal smoothness to produce a complete 3D pose. The second approach utilizes the spatial attention mechanism. ~\citet{handoccnet} propose a feature injection mechanism for occlusion-robust 3D hand mesh reconstruction. The third applies the amodal mask to perceive the invisible part. Amodal mask refers to the ability to perceive entire objects despite partial occlusion, which has the potential to make computers more human-like in handling occlusion. Ours is related to the amodal mask. Prior studies~\cite{zhan2020self, zhou2021human} have employed amodal masks to aid in recovering occluded 2D content from images. In our method, we leverage amodal masks to significantly enhance the optimization of neural implicit fields, thereby introducing a novel means of improving reconstruction in occluded regions. However, simply applying initial masks suffers from two issues: (1) some of them may be incorrect, and (2) they are not multi-view consistent. To address these issues, we use a semantic head to refine the masks, resulting in improved reconstruction quality.

\section{Methods}
Our objective is to reconstruct the 3D object from a video sequence $\{\image_\nframe\}_{k=0,...,N}$, where a hand holds a rigid object and rotates it in front of a static RGB camera. In our problem, hand poses are assumed to be fully constrained by objects. Therefore, hand poses are the same across different frames; only global translation and rotation of the hand may differ. 

We adopt the widely used 3D parametric model MANO~\cite{mano} to represent the hand. MANO can generate hand mesh by inputting two sets of parameters. Shape parameters $\shapeparameter \in \mathbb{R}^{10}$ control the hand shape and pose parameters $\poseparameter \in \mathbb{R}^{16\times 3}$ represent the rotation of 16 joints. We estimate the parameter of MANO along with the relative rotation $R\in SO(3)$ and translation $T \in \mathbb{R}^3$ between the hand and camera from RGB sequence images. Thus, the hand mesh can be defined as $\hand_\nframe=\{MANO(\shapeparameter, \poseparameter), R_\nframe, T_\nframe\}$, where $\nframe$ indicates the $\nframe_{th}$ frame, and $\shapeparameter, \poseparameter$ are shared for all the frames.  

The 3D object shape is represented by an SDF-based implicit function $\sdfnetwork$. By mapping a query 3D point to a signed distance from the object surface, we can extract the zero-level set as the object surface. As the object shape reconstruction is supervised by image sequences, we represent the object appearance by an extra implicit function $\colornetwork$. Both $\sdfnetwork$ and $\colornetwork$ are optimized through volume rendering to minimize differences between input images $\image_\nframe$ and rendered images $\hat{\image_\nframe}$. However, employing this technique alone leads to incomplete reconstruction results due to the absence of observations in the occluded region (hand-object contact area).

\par To address these challenges, we incorporate amodal masks and physical contact guidance into the neural rendering framework for constraining the reconstruction of the invisible parts. Specifically, an overview of our method is shown in \figref{fig_pipeline}. Given a monocular RGB input video of a moving hand-held object, our method reconstructs the hand-held object without any prior of the object category. Same with \cite{hhor}, we assume that the object is firmly grasped, enabling us to jointly predict the object motion by hand estimator. We apply the SDF-based implicit function to represent the object. To improve the reconstruction quality in contact areas, the 2D occlusion elucidation and the physical contact priors are leveraged. First, we utilize amodal masks to detect and supervise these regions. We ensure the consistency and quality of the amodal masks by refining them using an additional semantic head after the implicit neural network. Moreover, we apply contact constraints, which require that the object does not intersect with the hand and is in close proximity when they make contact.

\subsection{3D Hand Reconstruction}
\seclabel{sec_hand}
\par The first step of our framework is to perform hand pose estimation. We employ a learning-based approach to achieve a robust initialization and further optimize the hand model by fitting it to 2D keypoints.
\par Previous research~\cite{lv2021handtailor} has demonstrated that directly fitting MANO to 2D keypoints is highly non-linear and very sensitive to initial parameters. Therefore, we first utilize the pre-trained monocular hand reconstruction model HandOccNet~\cite{handoccnet} to estimate the hand model parameters of each frame and then average them to obtain a more robust initialization. The hand model can be optimized by minimizing the difference between 2D keypoints detections and reprojection of 3D joints:
\begin{equation}
    \hand_k^* = \mathop{\arg\min}\limits_{\hand_k}(\sum_{i=1}^{16}||\pi(\joint_{3d}^i(\hand_k))-\joint_{2d}^i||
    + \lambda_{reg}\loss_{reg}),
\end{equation}
where $\pi(.)$ denotes the projection operation, $\joint_{3d}$ and $\joint_{2d}$ represent the 3D and 2D joint location respectively. The last term $\loss_{reg}=||\shapeparameter||^2_2 + ||\poseparameter||^2_2$ is for regularization. We use Mediapipe~\cite{mediapipe} to obtain 2D keypoints.
\par Inspired by ~\cite{hasson2021towards}, we add an additional term into the optimization over $k$ frames to force temporal smoothness:
\begin{equation}
    \sum_\nframe ||\hand_\nframe - \hand_{\nframe-1}||.
\end{equation}
\par Jointly optimizing the energy function may be unstable. Therefore, we first optimize the relative rotation $R$ and translation $T$, followed by the optimization of the MANO parameters. After optimization, we can transfer RGB video sequences into multi-view images in the hand-centric coordinates, with the hand wrist serving as the origin.

\subsection{Object Reconstruction}
\seclabel{sec_obj}
\par A key to our framework is learning a hand-centric Signed Distance Function (SDF) representation, enabling the learning of a consistent 3D shape and appearance of the object. It is learned per sequence and does not require pre-training. To optimize the SDF, We adopt the NeuS method~\cite{neus} to utilize the volume rendering technique, while also integrating amodal mask and contact constraints.
\subsubsection{SDF-based Implicit Representation.}
\seclabel{sec_sdfl}
\par We represent the geometry and appearance by two MLP networks, a geometry network $\sdfnetwork:\mathbb{R}^3\rightarrow\mathbb{R}$ and a color network $\colornetwork:\mathbb{R}^3\times\mathbb{S}^2\rightarrow\mathbb{R}^3$. Given a 3D point $x$, the geometry network maps it to the SDF value $\sdfnetwork(x)$, and the color network takes $x$ along with view direction $\view$ as inputs and outputs color $\colornetwork(x, \view)$. The object surface is then extracted as the zero-level set of the SDF:
\begin{equation}
    \surface = \{x|\sdfnetwork(x) = 0\}.
\end{equation}
\par For each pixel, we sample a set of points along the corresponding camera ray, denoted as $\{\pixel_i=\origin+t_i\view|t_i\in[t_n, t_f]\}$, where $\pixel_i$ are the sampled points, $\origin$ is the camera position, $\view$ is the viewing direction, and $t_n, t_f$ denote the bound of the sample ray. Then we can get the rendered color as:
\begin{equation}
    \hat{c} = \sum_{i}T_i\alpha_i\colornetwork(\pixel_i, \view),
    \eqlabel{eq_render}
\end{equation}
where $T_i=\prod_{j=1}^{i-1}(1-\alpha_j)$ is the discrete accumulated transmittance, and $\alpha_i=1-exp(-\int_{t_i}^{t_{i+1}}\rho(t)dt)$ denotes the discrete opacity values. $\rho(t)$ is the opaque density transferred from SDF as defined in ~\cite{neus}.

\begin{figure}[!t]
  \centering
  \includegraphics[width=0.45\textwidth]{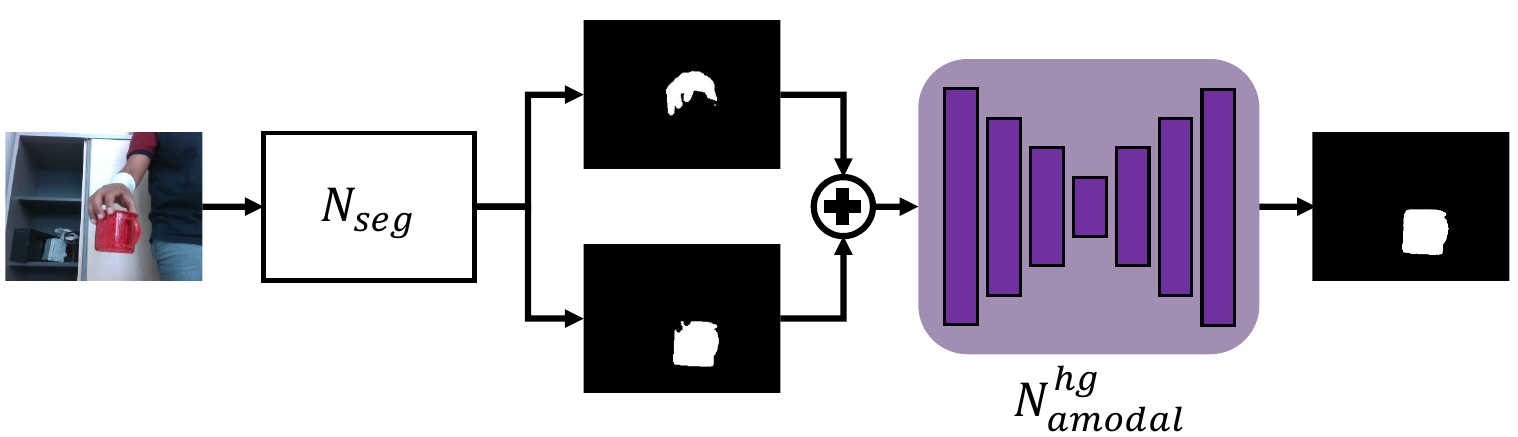}
  \caption{Overview of the amodal completion network architecture. Given initial segmentation maps of hand and object as input, an hourglass module is applied to produce the object amodal masks.}
  \figlabel{fig_amodal}
\end{figure}

\subsubsection{Amodal for Shape Completion.}
\seclabel{sec_amodal}
\par The amodal completion network targets at segmenting the invisible part of the object to offer the understanding of its complete shape, which can then be utilized to supervise the object geometry. Modern 2D amodal segmentation models~\cite{aisformer} trained on large labeled datasets can provide reasonable predictions. However, in our case, we require amodal masks for a range of categories that may not be present in the training dataset. Thus, we complete the hand and object segmentation maps into object amodal masks. This simultaneous input of hand and object segmentation maps is not restricted by object categories and effectively captures the patterns of hand-object interaction.

\par With this observation, we utilize a simple hourglass network~\cite{hourglass} $\amodalnetwork$ to estimate amodal masks $\hat{\amodal}$ ignoring the category information. Specifically, as \figref{fig_amodal} shows, we first obtain the segmentation maps of the hand and object from $\segnetwork$, an off-the-shelf method ~\cite{boerdijk2021s}. Then, the segmentation maps of hand and object $\seg$ are concatenated and fed into $\amodalnetwork$ to generate the amodal results. The network is trained on ObMan~\cite{obman}. ObMan is a large-scale hand-object interaction dataset, wherein ground-truth amodal masks $\amodal$ can be obtained by rendering 3D models. The cross-entropy loss $\loss_{CE}(.)$ is applied to supervise the predictions:
 \begin{equation}
     \loss_{amodal} = \loss_{CE}(\amodal, \hat{\amodal}).
 \end{equation}
\paragraph{Mask Refinement with View Consistency.}
\par Since the amodal mask of each frame is predicted independently, they lack multi-view consistency and are often inaccurate, especially when the object is heavily occluded, as shown in  ~\figref{fig_refine}(a). To resolve these inconsistencies and refine the masks, we use an additional semantic head. As demonstrated in ~\cite{zhi2021place}, the semantic neural field can naturally leverage the multi-view consensus to improve the accuracy of segmentation. Given a 3D point $x$, the semantic head predicts a logit $\logit(x)$, which is defined as:
\begin{equation}
     \logit(x) = \seghead(x),
\end{equation}
where $\seghead$ is also an MLP network. Similar to color, we adopt volume rendering to convert the semantic logits into 2D semantic maps denoted as $\renderlogit$, as presented in \Eqref{eq_render}:
\begin{equation}
    \hat{\renderlogit} = \sum_{i}T_i\alpha_i\seghead(\pixel_i).
\end{equation}
We then use a softmax to compute the probabilities and supervise by predicted amodal masks $\hat{\amodal}$ using the classification loss:
\begin{equation}
    \loss_{seg} = \loss_{CE}(\hat{\amodal}, \hat{\renderlogit}).
\end{equation}
\par The semantic head is trained together with the implicit neural network. After several iterations, we obtain refined amodal masks $\hat{\amodal}_{refined}$ by thresholding the probabilities of $\hat{\renderlogit}$, which are then used to supervise the geometry again. ~\figref{fig_refine}(b) presents examples of refined masks, which demonstrate the effectiveness in improving the accuracy and consistency of masks.

\begin{figure}[!t]
  \centering
  \includegraphics[width=0.45\textwidth]{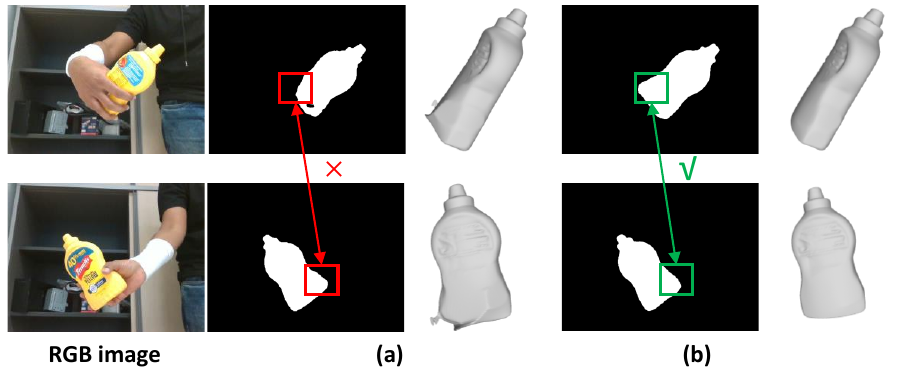}
  \caption{Qualitative results of mask refinement and reconstructed results. (a) Predicted masks from the completion network and corresponding mesh. (b) Refined results and corresponding meshes.}
  \figlabel{fig_refine}
\end{figure}

\subsubsection{Hand-Object Contact Constraints.}
\seclabel{sec_constraint}
\par We leverage the constraints that guide objects interacting in physical contact. In particular, when grasping the objects, there is no interpenetration between the hand and the object, and that, contacts occur at the surface of both. We express these contact constraints as a differentiable loss $\loss_{contact}$, that can be easily applied in the neural rendering framework.
\paragraph{Penetration.} To prevent penetration between the hand and object, we define a penetration loss $\loss_P$. Following ~\cite{ye2022s}, we penalize if any hand mesh vert $\vertices$ is predicted to have negative SDF value by geometry network $\sdfnetwork$, which can be formulated as:
\begin{equation}
    \loss_P = \sum_{\vertices \in \hand}||max(-\sdfnetwork(\vertices), 0)||.
\end{equation}
\paragraph{Attraction.} We further define an attraction loss $\loss_A$ to encourage the contact. 
We sample the rays only from pixels with the amodal mask value of $1$. By calculating the surface intersection of these rays, we can obtain the object surface point $\surfacepoint$, along with its corresponding normal $\surfacenormal$ in the contact area. Then we cast a ray along $\surfacenormal$ direction and find the nearest point it intersects the hand mesh. We determine whether the object is in contact with the hand based on the distance $\distance$ between the surface point $\surfacepoint$ and the intersection point $\handsurfacepoint$. The process is illustrated in ~\figref{fig_pipeline}. For object surface points in contact with hands (the distance $\distance$ smaller than a threshold $ \yuzhi$), we first encourage the object surface to be close to the hand surface. This involves ensuring that $\sdfnetwork(\handsurfacepoint)$ approaches 0, denoted as $\loss_A = ||\sdfnetwork(\handsurfacepoint)||$. However, we find this constraint for points in contact only usually does not reconstruct the surface we desire, as shown in \figref{fig_cs}. This can be attributed to: 1) objects surface points near the hand but not in contact are usually occluded in all images, lacking constraints for the reconstruction; 2) the hard threshold results in abrupt surface constraint changes near contact and non-contact regions; 3) the hard threshold is sensitive to the hand reconstruction quality. Consequently, we also introduce constraints for object surface points near the contact regions but not in contact by encouraging the SDF values to be a function of the distance between a surface point and the hand surface.

Therefore, our overall attraction loss $\loss_A$ is defined as :
\begin{equation}
    \loss_A = \begin{cases}
    ||\sdfnetwork(\handsurfacepoint)||& \distance < \yuzhi, \\
    ||\sdfnetwork(\handsurfacepoint) - \tanhxishu tanh(\frac{\distance-\yuzhi}{\tanhxishu})||& \distance \geq  \yuzhi,
\end{cases}
\end{equation}
where $\yuzhi, \tanhxishu$ are the hyper-parameters.  In our study, we empirically set $\yuzhi=0.001, \tanhxishu=0.5$.
\par Finally, we employ the surface smooth regularization~\cite{unisurf} in the contact region, which encourages $\surfacenormal$ to be similar with its neighborhood:
\begin{equation}
    \loss_S = \sum ||\surfacenormal-\surfacenormalneig||_2.
\end{equation}
\par Our final contact loss can be formulated as:
\begin{equation}
    \loss_{contact} = \xishu_P\loss_P + \xishu_A\loss_A + \xishu_S\loss_S.
\end{equation}
\subsection{Training}
\par In the training stage, we employ multiple loss functions to optimize the neural implicit field. Specifically, during training, we sample $\raynum$ rays and their corresponding reference colors $\pixelcolor_i$, and amodal mask values $\hat{\amodal}_i$. We use the amodal masks obtained by mask refinement as ground truth for the corresponding calculation. For each ray, we sample $\pointnum$ points.
The color loss $\loss_{color}$ is defined as:
\begin{equation}
    \loss_{color} = \frac{1}{\raynum}\sum_i||\hat{\pixelcolor_i} - \pixelcolor_i||.
\end{equation}
We apply Eikonal loss~\cite{gropp2020implicit} and mask loss to regularize the SDF:
\begin{equation}
    \loss_{eik} = \frac{1}{\raynum\pointnum}\sum_{k,i}(||\nabla\sdfnetwork(\pixel_{k, i})||_2-1)^2,
\end{equation}
\begin{equation}
    \loss_{mask} = \loss_{CE}(\hat{\weight_i}, \hat{\amodal}_i),
\end{equation}
where $\hat{\weight_i}=\sum_{k}T_k\alpha_{i,k}$ is the sum of weight along the ray.
The overall training loss is:
\begin{equation}
\begin{split}
    \loss = \loss_{color} + \xishu_{mask}\loss_{mask} + \xishu_{eik}\loss_{eik} + \\ \xishu_{seg}\loss_{seg} + \xishu_{contact}\loss_{contact},
\end{split}
\end{equation}
where $\xishu_{mask}=10, \xishu_{eik}=0.1, \xishu_{seg}=0.1, \xishu_{contact}=5$ are set empirically.
\paragraph{Optimize Camera Poses.} The estimated camera pose from the hand pose may not be accurate due to occlusion between the hand and object, leading to a significant degradation in the quality of the reconstruction. Pose refinement has been explored in previous NeRF-based models~\cite{barf}. We incorporate this to effectively optimize for the poses jointly with the object representation.

\begin{figure}[!t]
  \centering
  \includegraphics[width=0.45\textwidth]{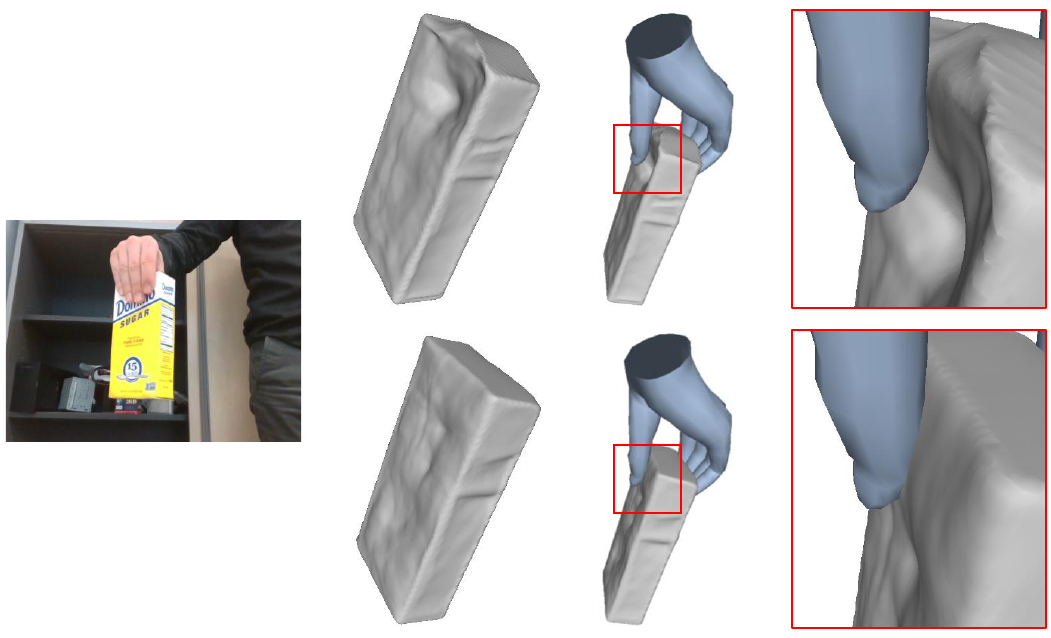}
  \caption{Comparison of different contact loss designs. Top: reconstructed mesh using the contact loss only constraining the object surface points in contact. Bottom: reconstructed mesh using our loss.}
  \figlabel{fig_cs}
\end{figure}

\begin{figure}[!t]
  \centering
  \includegraphics[width=0.46\textwidth]{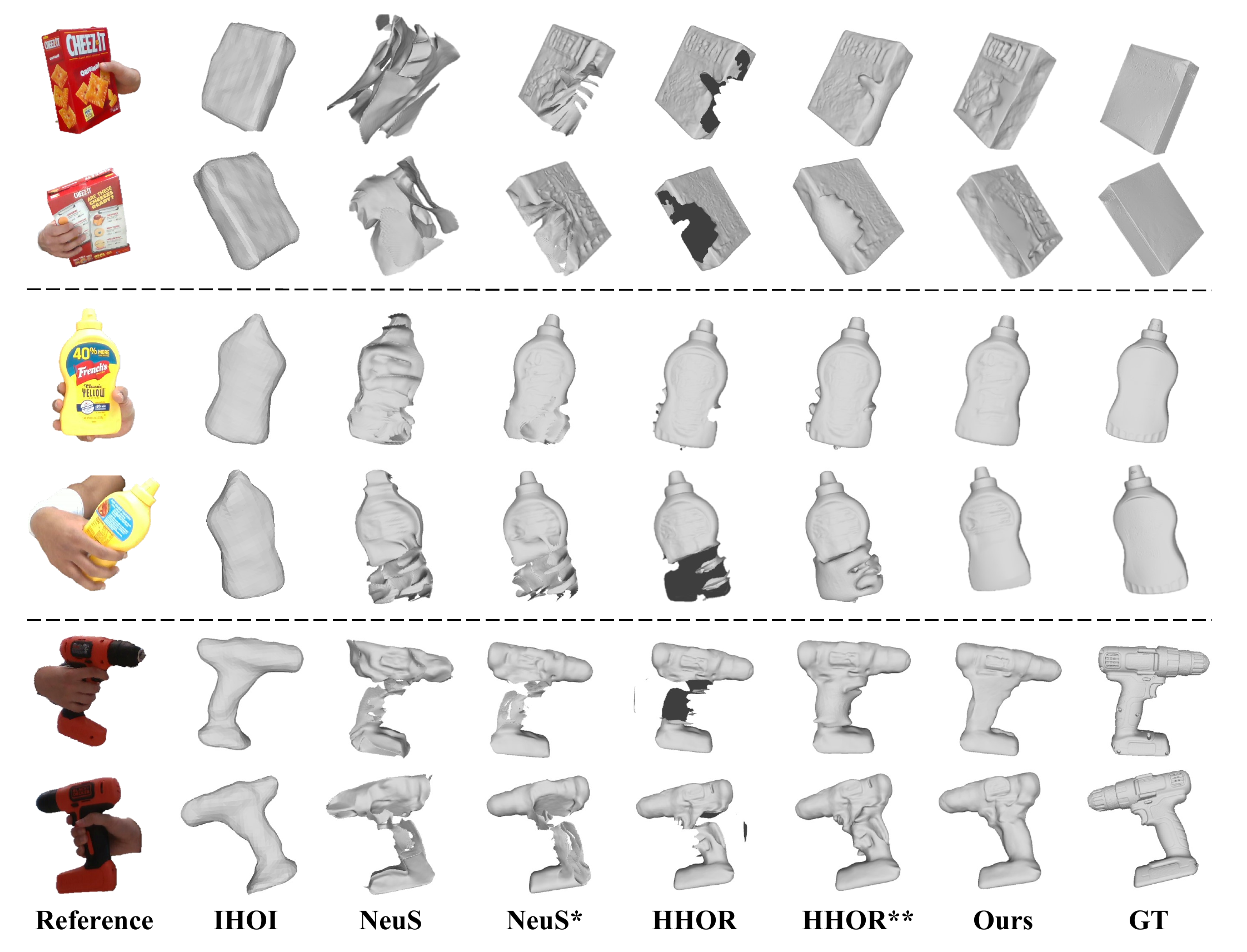}
  \caption{Qualitative comparison with baseline methods on the HO3D dataset. $^*$ indicates that the method uses the gound-truth camera pose. $^{**}$ indicates that the method with post-processing. Compared with other methods, we can produce more complete and detailed reconstruction results. Zoom in for details.}
  \figlabel{fig_ho3dres}
\end{figure}

\begin{figure}[!t]
  \centering
  \includegraphics[width=0.45\textwidth]{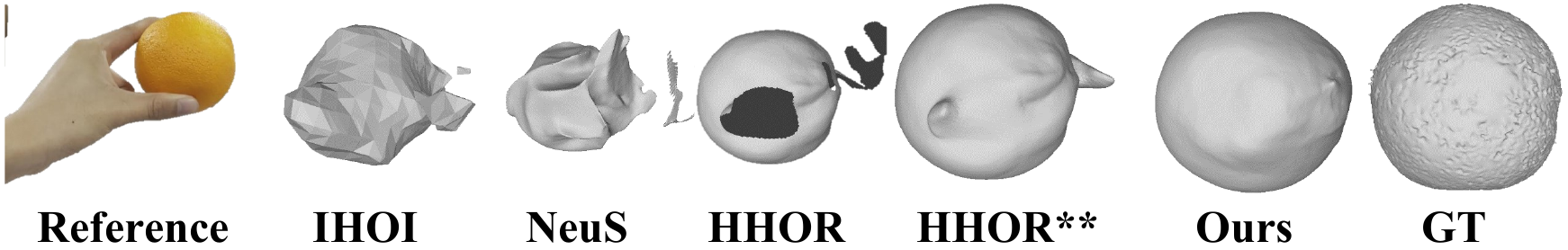}
  \caption{Qualitative comparison with baselines methods on HOD dataset.} 
  \figlabel{fig_hodres}
\end{figure}
\section{Experiments}
\par In this section, we first present the hand-object interaction datasets and evaluation metrics. Subsequently, we compare our method to state-of-the-art approaches and provide ablation results. 
\subsection{Experimental Setups}
\paragraph{Implementation Details.} We use the same network architecture as NeuS~\cite{neus}, following them to normalize all cameras within a unit sphere and initialize network parameters to approximate the SDF to a unit sphere. For training the model, we use Adam optimizer~\cite{adam} with a learning rate of $5e-4$ and sampled 1024 rays per batch for a total of $100k$ iterations. The training takes about 14 hours in total on a single NVIDIA RTX3090 GPU. Our implementation is based on PyTorch.
\paragraph{Datasets.} To evaluate our method, we perform the experiments on HO3D~\cite{ho3d} and HOD~\cite{hhor}. 
\begin{itemize}
\item HO3D is a dataset that contains the RGBD video of a hand interacting with YCB objects~\cite{ycb} with 3D annotations of both hand and object. We select the 5 sequences in which the objects are firmly grasped by users for our experiments. 
\item HOD aims to reconstruct hand-held objects from RGB sequences, containing 35 objects. However, only 14 ground truth scanned meshes are available for evaluation.
\end{itemize}
In the experiment, we use 500 frames from each sequence of HO3D and all provided frames of HOD.
\paragraph{Evaluation Metrics.} We evaluate both the quality of object reconstruction and the relationship between the hand and object. Firstly, we use Marching Cubes~\cite{lorensen1987marching} to extract the object mesh from SDF. Following prior research, we then evaluate the object reconstruction quality using \textbf{Chamfer Distance (CD)}. As the reconstructed result and ground truth mesh are in different coordinates, we follow~\citet{hhor} to normalize each mesh to unit size and apply the ICP to register the reconstructed mesh with the ground truth mesh. For evaluating the relationship between the hand and object, we report the \textbf{Intersection Volume (Vol)} in $cm^3$ between the hand mesh and object mesh, similar to ~\cite{ye2022s}.
\paragraph{Comparison Baselines.} In our evaluation, we compare our method with several existing approaches, including (1) \textbf{HHOR}~\cite{hhor}, which addresses the same problem as ours. Additionally, we present the results of HHOR with post-processing (denoted as \textbf{HHOR$^{**}$}) using MeshLab~\cite{cignoni2008meshlab} to remove unnecessary parts and fill holes; (2) \textbf{NeuS}~\cite{neus} serves as the foundation for our method. Since the reconstruction quality of NeuS is greatly influenced by the accuracy of the camera pose, we also report its results on HO3D using ground truth camera poses (denoted as \textbf{NeuS$^{*}$}) for a fair comparison, as HOD does not offer ground truth data; (3) \textbf{IHOI}~\cite{ye2022s}, a learning-based single image hand-held object reconstruction method that is pre-trained on sequences of the HO3D and other datasets. We evaluate IHOI on each frame of the sequence and report the average results. As the results of NeuS and HHOR are not watertight, we do not report the intersection volume metric.

\begin{table}[!t]
  \centering
  \begin{tabular}{LLLLL}
   \toprule
   \multicolumn{1}{c}{\multirow{2}{*}{Methods}} &\multicolumn{2}{c}{HO3D} &\multicolumn{2}{c}{HOD}\\
    \cmidrule(r){2-5} 
    &CD$\downarrow$ &Vol$\downarrow$ & CD$\downarrow$ &Vol$\downarrow$ \\
    \midrule
    IHOI&2.206 &2.192 &6.607 &\textbf{0.505}  \\
    NeuS&3.310 &- &3.093 &-\\
    NeuS$^*$&0.872 &- &- &- \\
    HHOR&1.256 &- &0.589 &-  \\
    HHOR$^{**}$&0.591 &7.771 &0.347 &1.738 \\
    Ours&\textbf{0.282} &\textbf{0.327} &\textbf{0.277} &0.757 \\
    \bottomrule
    \end{tabular}
    \caption{Quantitative results of object reconstruction on the HO3D and HOD datasets using Chamfer distance (in unit size) and intersection volume ($cm^3$). $^*$ indicates that the method uses the ground-truth camera pose. $^{**}$ indicates that the method with post-processing.}
    \tablelabel{table1}
\end{table}
\subsection{Comparisons With the State-of-the-Art Methods}
\par We evaluate reconstructed 3D meshes on HO3D and HOD. Averaged quantitative results are presented in \tableref{table1}. Please refer to the Supp. for more detailed results.
\paragraph{Comparison Results on HO3D.} We visualize the reconstructed objects in ~\figref{fig_ho3dres}. The learning-based method IHOI can predict the coarse shape of the object, but it typically loses the finer details of the object surface when compared to neural rendering methods. Inaccurate camera poses significantly decrease the reconstruction quality of NeuS, but when ground-truth poses are used  (NeuS*), it achieves similar reconstruction quality to HHOR in the visible part of the object. However, both NeuS and HHOR struggle to handle occlusion, which leads to incomplete surface reconstructions. While HHOR$^{**}$ (HHOR with post-processing) can use Poisson Reconstruction to aid in filling the holes, the resulting object reconstruction may contain obvious artifacts. This is because the Poisson Reconstruction can not correctly fill the surface for the missing part when a large part of the object is occluded. In contrast, our method can recover detailed object meshes in both the visible and invisible parts without any post-processing.
\par By analyzing quantitative results on HO3D, our method significantly outperforms the comparison methods in both 3D reconstruction and hand-object relationships. Other volume rendering-based methods, NeuS with ground-truth poses and HHOR achieve better performance of reconstruction than the learning-based method IHOI. However, they still struggle in reconstructing complete geometry. When HHOR$^{**}$ can obtain complete surfaces with $0.591$ CD, our approach achieves even lower $0.282$ CD values, demonstrating an improvement of $52\%$. In terms of intersection volume, our approach outperforms IHOI and significantly surpasses HHOR$^{**}$. This can be attributed to the evident hand artifacts in HHOR$^{**}$ that lead to increased volume, highlighting the effectiveness of our integrated contact constraints.
\paragraph{Comparison Results on HOD.} HOD contains objects with more complex shapes and textures, but less hand occlusion. For reconstruction quality, our method continues to surpass the state-of-the-art methods. Compared to HHOR$^{**}$, our method improves by $20\%$. Furthermore, visualizations in ~\figref{fig_hodres} highlight that other results still contain obvious artifacts, whereas our outcomes appear more reasonable. Our method can reconstruct a complete and detailed object mesh regardless of whether the hand-grasping type involves weak or heavy occlusion. Note that the learning-based method heavily relies on the learned prior, and therefore does not work well for objects beyond the training dataset. They cannot recover the shape accurately. Regarding the hand-object relationship, our method outperforms HHOR$^{**}$, emphasizing contact constraints' importance in less occluded scenarios. Though IHOI results in lower intersection volume, their predicted object shapes are absolutely inaccurate. Conversely, our method can reconstruct detailed object meshes with a reasonable hand-object relationship. 
\begin{figure}[!t]
  \centering
  \includegraphics[width=0.45\textwidth]{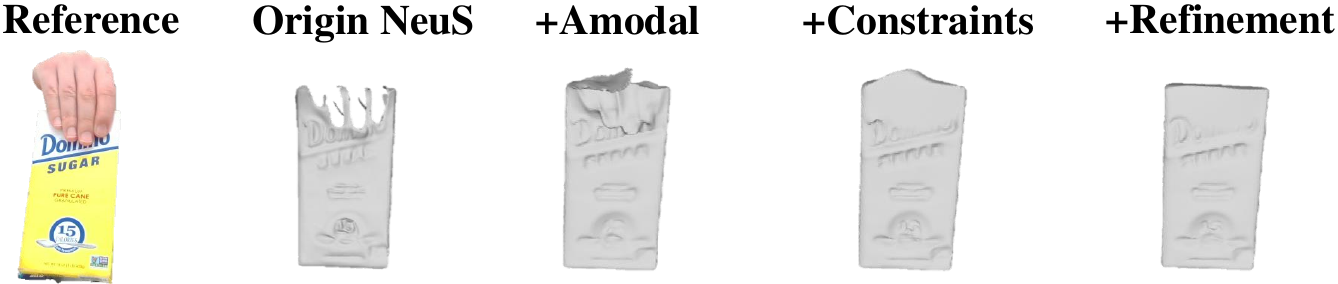}
  \caption{Qualitative results of ablation study. +Amodal indicates the utilization of amodal mask supervision, +Constraints indicates the incorporation of $\loss_{contact}$, and +Refinement indicates the application of mask refinement.}
  \figlabel{fig_ablation}
\end{figure}
\begin{table}[!t]
  \centering
  \begin{tabular}{LLLLL}
   \toprule
    NeuS & Amodal &Constraints &Refinement &CD$\downarrow$ \\
    \midrule
    \CheckmarkBold& & & &1.094\\
    \CheckmarkBold&\CheckmarkBold & & &0.448\\
    \CheckmarkBold&\CheckmarkBold &\CheckmarkBold & &0.333\\
    \CheckmarkBold&\CheckmarkBold &\CheckmarkBold &\CheckmarkBold &0.282\\
    \bottomrule
    \end{tabular}
    \caption{Ablation studies of each component of our method over 5 objects on the HO3D dataset. We report the averaged 3D object reconstruction metric.}
    \tablelabel{table2}
\end{table}
\begin{table}[!t]
    \centering
    \begin{tabular}{LLL}
    \toprule
    Methods&CD$\downarrow$ &Vol$\downarrow$\\
    \midrule
    Ours w/o $\loss_A$& 0.376& \textbf{0.108}\\
    Ours w/o $\loss_P$& 0.319& 0.614\\
    Ours w/ ${\loss_A}^-$&0.341 &0.439 \\
    Ours&\textbf{0.282}&0.327\\
    \bottomrule
    \end{tabular}
    \caption{Ablation on $\loss_{contact}$ variants.}
    \tablelabel{table_cs}
\end{table}

\subsection{Ablation Studies}
\par To evaluate the effectiveness of our proposed components, we perform experiments on HO3D across four distinct settings: (1) NeuS; (2) NeuS with amodal masks; (3) NeuS with amodal masks and contact constraints; (4) Ours: NeuS with amodal masks, contact constraints, and mask refinement. 
\par According to \tableref{table2}, incorporating amodal masks significantly improves reconstruction quality, reducing the CD value by $0.646$. Visualizations in \figref{fig_ablation} demonstrate successful recovery of overall shape, indicating that amodal masks effectively fuse observations for complete reconstruction.
\par With contact constraints, we can further improve the geometry quality with $0.115$ CD value decrement. As demonstrated in \figref{fig_ablation}, the utilization of $\loss_{contact}$ effectively reduces the ambiguities caused by occlusion in the contact region, resulting in a smoother surface.
\par Moreover adding mask refinement can remove the wrongly estimated part caused by inaccurate amodal masks, leading to a $0.051$ CD value reduction. The visualization results illustrate that the mask refinement effectively removes wrong results at the object boundaries. These results demonstrate the effectiveness of our proposed components.
\subsubsection{Different $\loss_{contact}$ Design.}
\par We analyze the results with various $\loss_{contact}$ designs in \tableref{table_cs}. We conduct experiments without attraction or penetration loss. Incorporating only penetration loss minimizes intersection volume, yet its reconstruction quality lags behind other methods. Conversely, solely applying attraction loss increases intersection volume while enhancing reconstruction. To balance reconstruction quality and intersection volume, we simultaneously apply these two losses in our method. Furthermore, we conducted a comparison by substituting our formulated $\loss_A$ with constraints on object surface points in contact only (denoted as ${\loss_A}^-$). Our approach reaches lower CD values and intersection volume, demonstrating the efficacy of guidance in the vicinity of the hand but not in the contact area.
\section{Conclusions and Future Work}
\par In this paper, we present a framework for reconstructing the 3D generic objects in hand using a monocular RGB video. The key insights of our method are to incorporate the amodal masks and physical contact guidance for dealing with surface reconstruction under occlusion. On several datasets, we have demonstrated state-of-the-art results compared with existing methods. In the future, we aim to speed up the training process by integrating hybrid neural representations such as ~\cite{muller2022instant}, and relax the assumption of fixed grasping by inferring the object pose~\cite{wen2023bundlesdf}.
\section*{Acknowledgments}
\par This work was supported in part by NSFC under Grants (62103372, 62088101, 62233013), the Fundamental Research Funds for the Central Universities (226-2023-00111), and the OPPO Research Fund.
\bibliography{aaai24}
\newpage
\begin{appendices}
\part*{\centering Supplementary Material} 
\addcontentsline{toc}{part}{Supplementary Material} 
\begin{figure*}[!ht]
  \centering
  \includegraphics[width=0.85\textwidth,trim={0.3cm 1.5cm 1.2cm 1.5cm},clip]{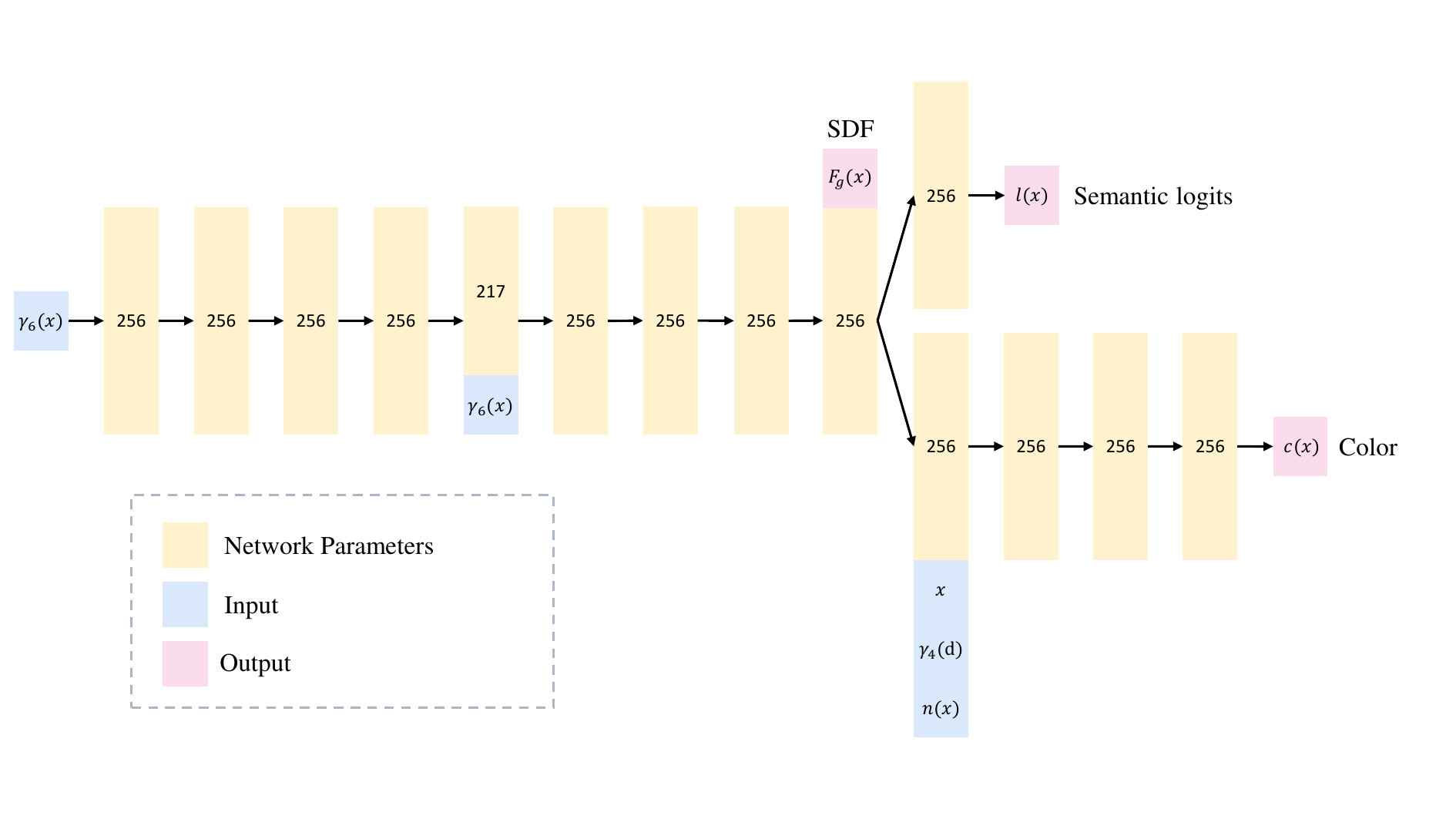}
  \caption{Implicit Neural Network Architecture.}
  \figlabel{fig_network}
\end{figure*}
\section{Network Architecture}
\par We present the implicit neural network architecture in ~\figref{fig_network}. We use a similar network architecture as NeuS, which consists of three MLPs to encode SDF $\sdfnetwork(x)$, semantic logits $\logit(x)$, and color $c(x)$. We use positional encoding for the input spatial point $x$ and view direction $d$. The SDF function consists of eight hidden layers, with a skip connection linking the input to the fourth layer's output. For the Semantic function, we added an extra layer to compute the logit. As for the color function, it has four hidden layers and takes the spatial point $x$, view direction $d$, normal vector $n(x)$, and feature vector from the SDF function output as inputs.


\section{Quantitative Results on Individual Objects}
\par We conduct reconstruction experiments on overall 19 objects from HO3D and HOD datasets, and compare our method with SOTA methods. We show quantitative results on each individual object in ~\tableref{table_detail}.

\section{Robustness Against Hand Prediction Quality}
\par We utilize the provided ground truth hand poses of HO3D for an assessment of our method's robustness. As depicted in ~\tableref{table_hand}, our predictions yield comparable results to those derived from the ground truth. This achievement can be primarily attributed to the high accuracy of our predicted hand poses, further enhanced by simultaneous optimization, leading to improved motion precision.

\begin{table*}[!t]
  \centering
  \begin{tabular}{llllllllllllll}
   \toprule
   \multicolumn{2}{c}{Objects} &\multicolumn{2}{c}{IHOI} &\multicolumn{2}{c}{NeuS} &\multicolumn{2}{c}{NeuS$^*$} &\multicolumn{2}{c}{HHOR} &\multicolumn{2}{c}{HHOR$^{**}$} &\multicolumn{2}{c}{Ours} \\
    \cmidrule(r){3-14} 
    & &CD$\downarrow$ &Vol$\downarrow$ & CD$\downarrow$ &Vol$\downarrow$ & CD$\downarrow$ &Vol$\downarrow$ & CD$\downarrow$ &Vol$\downarrow$ & CD$\downarrow$ &Vol$\downarrow$ & CD$\downarrow$ &Vol$\downarrow$ \\
    \midrule
    &Cracker Box &2.242 &1.240 &3.845 &- &0.931 &- &0.956 &- &0.146 &2.766 &\textbf{0.084}&\textbf{0.396} \\
    &Sugar Box &1.053 &1.618 &2.695 &- &0.567 &- &1.335 &- &0.438 &11.113 &\textbf{0.227}&\textbf{0.069}\\
    HO3D&Power Drill &2.774 &4.328 &4.780 &- &0.830 &- &1.389 &- &0.743 &4.245 &\textbf{0.321}&\textbf{0.228} \\
    &Mustard Bottle &0.386 &2.824 &1.322 &- &0.566 &- &0.627 &- &0.281 &14.814 &\textbf{0.130}&\textbf{0.703} \\
    &Mug &4.577 &0.949 &3.907 &- &1.466 &- &1.974 &- &1.348 &5.915 &\textbf{0.646}&\textbf{0.237} \\
    \midrule
     &Mean &2.206 &2.192 &3.310 &- &0.872 &- &1.256 &- &0.591 &7.771 &\textbf{0.282}&\textbf{0.327}\\
    \midrule
    &Orange &11.769 &0.497 &5.763 &- &- &- &0.679 &- &\textbf{0.419} &1.446 &0.424 &\textbf{0.454} \\
    &Duck &4.985 &0.430 &3.941 &- &- &- &0.984 &- &0.681 &\textbf{0.110} &\textbf{0.507} &0.889 \\
    &Robot &2.702 &\textbf{0.298} &1.305 &- &- &- &0.328 &- &0.275 &0.544 &\textbf{0.187} &0.503 \\
    &Cat &26.926 &0.967 &6.008 &- &- &- &0.614 &- &0.384 &2.584 &\textbf{0.125}&\textbf{0.625} \\
    &Pingpong &5.621  &\textbf{0.451} &4.707 &- &- &- &0.765 &- &0.608 &1.212 &\textbf{0.404} &1.093 \\
    &Box &3.995 &\textbf{0.272} & 1.706 & - & - & - & 0.841 & - & 0.636 & 0.644 & \textbf{0.387} & 1.127 \\
    HOD&AirPods &7.072 & \textbf{0.102} & 1.329 & - & - & - & 0.794 & - & \textbf{0.092} & 0.245 & 0.103 & 0.119 \\
    &Bottle &2.995 & \textbf{0.567} & 0.997 & - & - & - & 0.855 & - & 0.536 & 1.821 & \textbf{0.481} & 1.155 \\
    &Case &6.015 & \textbf{0.096} & 1.753 & - & - & - & 1.081 & - & \textbf{0.344} & 0.186 & 0.386 & 0.181 \\
    &Apollo &4.415 & \textbf{0.628} & 6.014 & - & - & - & 0.462 & - & 0.253 & 3.986 &\textbf{0.210} & 1.062 \\
    &David &3.139 & 0.894 & 3.013 & - & - & - & 0.231 & - & 0.208 & 2.352 & \textbf{0.195} & \textbf{0.892} \\
    &Giuliano &4.087 & \textbf{0.143} & 0.599 & - & - & - & 0.205 & - & \textbf{0.107} & 2.214 & 0.171 & 0.687 \\
    &Marseille &4.654 & 0.981 & 2.207 & - & - & - & 0.192 & - & 0.181 & 3.819 & \textbf{0.162} & \textbf{0.838} \\
    &Moliere &4.122 & \textbf{0.738} & 3.965 & - & - & - & 0.218 & - & 0.137 & 3.172 & \textbf{0.136} & 0.979 \\
    \midrule
    &Mean &6.607 &\textbf{0.505} &3.093 &- &- &- &0.589 &- &0.347 &1.738 &\textbf{0.277} &0.757 \\
    \bottomrule
    \end{tabular}
    \caption{Individual quantitative results of object reconstruction on the HO3D and HOD datasets. $^*$ indicates that the method uses the gound-truth camera pose. $^{**}$ indicates that the method with post-processing.}
    \tablelabel{table_detail}
\end{table*}

\section{Results for Amodal Completion}
\par We evaluate the amodal completion on the HO3D dataset, as shown in ~\tableref{table_amodal} and ~\figref{fig_amodal1}. To evaluate the quality of the completed masks, we adopt Intersection over Union (IoU) as our metrics. Our completion network can achieve high mIOU values, while the refinement further improves the results. Moreover, ~\figref{fig_amodal2} demonstrates that our method is capable of accurately obtaining amodal masks for objects with complex shapes in the HOD datasets.


\begin{table*}[!t]
  \begin{minipage}[t]{0.5\linewidth}
    \centering
    \begin{tabular}{lllll}
    \toprule
    Objects &\multicolumn{2}{c}{Prediction} &\multicolumn{2}{c}{Ground Truth}\\
    \cmidrule(r){2-5} 
    &CD$\downarrow$ &Vol$\downarrow$ &CD$\downarrow$ &Vol$\downarrow$\\
    \midrule
    Cracker Box&0.084&0.396 &0.081 &0.307\\
    Drill&0.321 &0.228 &0.298 &0.193\\
    Sugar Box&0.227 &0.069 &0.186 &0.077\\
    Bottle&0.130 &0.703 &0.103 &0.688\\
    Mug&0.646 &0.237 &0.607 &0.359\\
    \bottomrule
    \end{tabular}
    \captionsetup{justification=raggedright}
    \caption{Analysis of robustness in relation to hand prediction quality.}
    \tablelabel{table_hand}
  \end{minipage}%
  \begin{minipage}[t]{0.5\linewidth}
    \centering
    \begin{tabular}{lll}
    \toprule
    Objects & Raw &+Refinement\\
    \midrule
    Cracker Box&98.15 &\textbf{99.02}\\
    Drill&93.15 &\textbf{94.37}\\
    Sugar Box&92.84 &\textbf{94.28}\\
    Bottle&92.51 &\textbf{95.17}\\
    Mug&95.29 &\textbf{96.52}\\
    \bottomrule
    \end{tabular}
    \captionsetup{justification=raggedright}
    \caption{Quantitative results of amodal completion on the HO3D dataset. We report the mIOU\%.}
    \tablelabel{table_amodal}
  \end{minipage}
\end{table*}

\begin{table*}[!t]
  \begin{minipage}[t]{0.5\linewidth}
    \centering
    \begin{tabular}{lll}
    \toprule
    Objects & Origin &+Pose Refine\\
    \midrule
    Cracker Box&0.431 &\textbf{0.427}\\
    Drill&0.208 &\textbf{0.187}\\
    Sugar Box&0.208 &\textbf{0.202}\\
    Bottle&0.127 &\textbf{0.091}\\
    Mug&0.225 &\textbf{0.200}\\
    \bottomrule
    \end{tabular}
    \captionsetup{justification=raggedright}
    \caption{Quantitative results of camera-relative motion on the HO3D dataset.}
    \tablelabel{table_cam}
  \end{minipage}%
  \begin{minipage}[t]{0.5\linewidth}
    \centering
    \begin{tabular}{lll}
    \toprule
    Objects & PA-MPJPE$\downarrow$ &PA-MPVPE$\downarrow$\\
    \midrule
    Cracker Box&4.704 &4.184\\
    Drill&3.891 &3.265\\
    Sugar Box&3.113 &3.485\\
    Bottle&3.044 &3.763\\
    Mug&4.971 &4.855\\
    \bottomrule
    \end{tabular}
    \caption{Quantitative results of hand pose on the HO3D dataset.}
    \tablelabel{table3}
  \end{minipage}
\end{table*}



\section{Results for Hand Reconstruction}
\par We assess the quality of our hand reconstruction results on 5 sequences from the HO3D dataset.

\subsection{Camera-relative Motion Evaluation}
\par We evaluate the relative rotation $R$ and translation $T$ between the hand and camera. Since the 3D model and poses are retrieved up to a 3D similarity transformation, we apply Umeyama method~\cite{umeyama1991least} to align the estimated motion with the ground truth and calculate the Absolute Trajectory Error (ATE) using evo tools~\cite{evo}. 
\paragraph{Absolute Trajectory Error} The ATE is a widely adopted metric in SLAM used for investigating the global consistency of a trajectory. We denote the pose as $P=[R|T]$. ATE is based on the absolute relative pose between two poses $P_{ref,i}, P_{est,i} \in SE(3)$ at frame index $i$:
\begin{equation}
    E_i=P_{ref,i}^{-1} P_{est,i}.
\end{equation}
We use the full relative pose $E_i$ for calculation:
\begin{equation}
    ATE_i = ||E_i-I_{4\times4}||_{F}
\end{equation}
\par We report the mean value of 500 frames, comprising both the originally predicted results and the results obtained after pose refinement. The results are shown in ~\tableref{table_cam}, ~\figref{fig_motion}, and ~\figref{fig_motion2}, which demonstrate that simultaneous optimization of camera poses and radiance field improves the pose accuracy.

\subsection{Hand Pose Evaluation}
\par We evaluate the hand pose $MANO(\theta, \beta)$. We compute the PA-MPJPE and PA-MPVPE between the predicted hand mesh and the ground truth. The PA-MPJPE quantifies the average per-joint position error in Euclidean distance (mm), where PA-MPVPE measures the difference of the vertices. The quantitative results are shown in ~\tableref{table3}. Our method can accurately reconstruct hand poses by fusing video observations.

\begin{figure*}[!t]
  \centering
  \includegraphics[width=\textwidth,trim={1cm 2.5cm 1cm 1.5cm},clip]{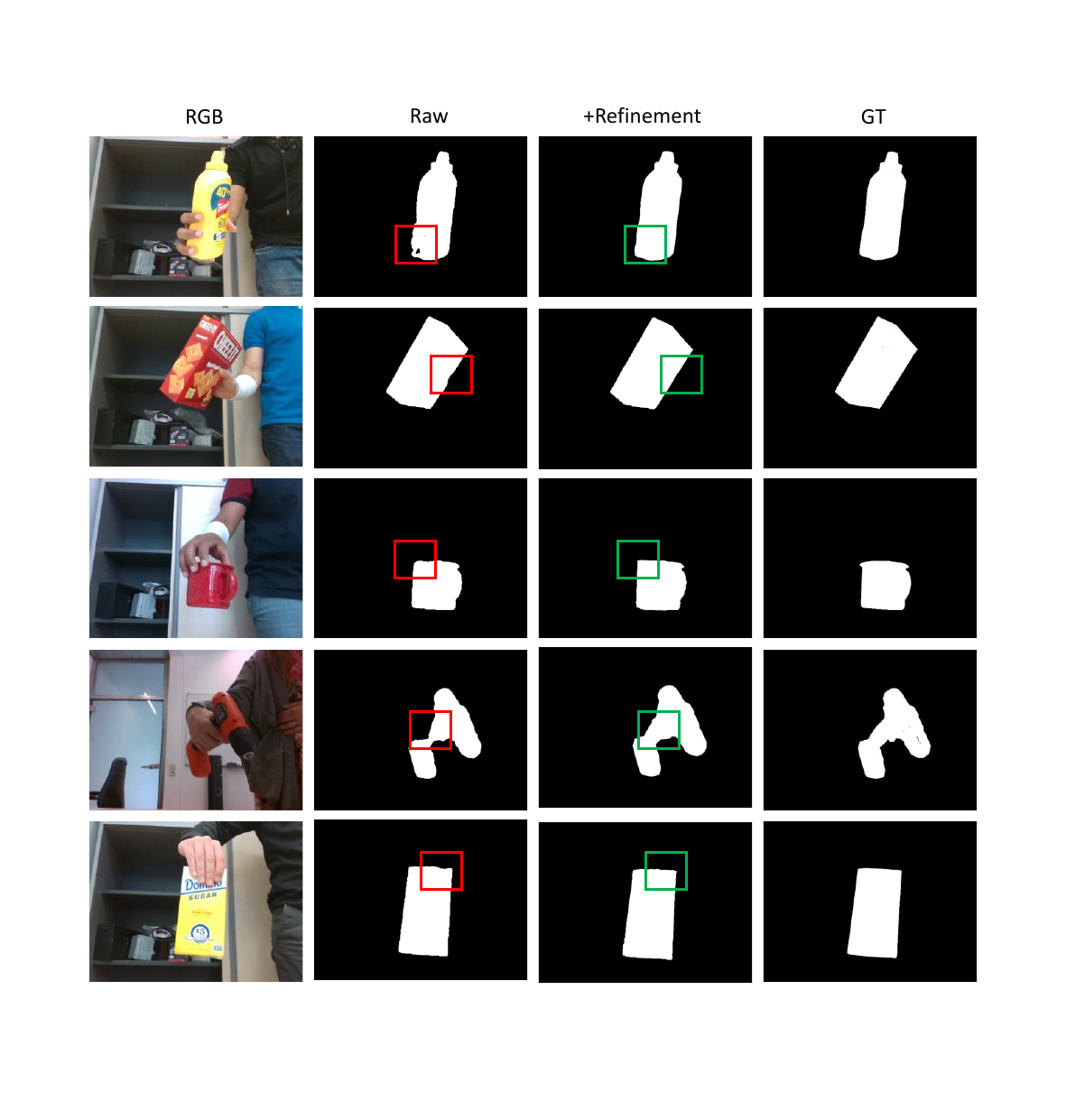}
  \caption{Qualitative amodal completion results on the HO3D dataset. Raw corresponds to the amodal masks predicted by the completion network, and +Refinement represents the results after the mask refinement. Our completion network can obtain great amodal masks, while the refinement further improves the results, as shown in the highlighted rectangle area.}
  \figlabel{fig_amodal1}
\end{figure*}
\begin{figure*}[!t]
  \centering
  \includegraphics[width=0.8\textwidth,trim={2cm 1cm 2cm 0.5cm},clip]{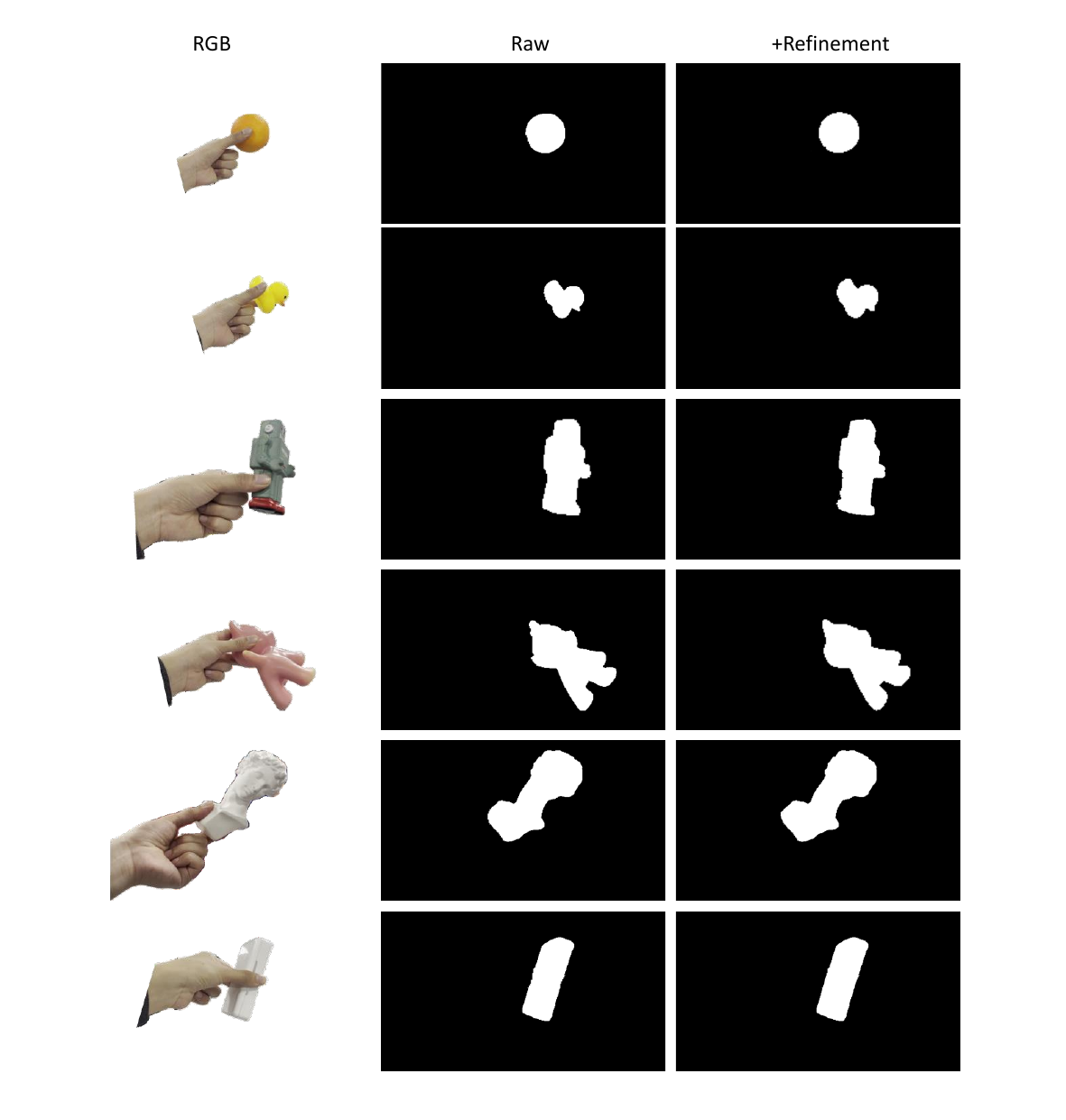}
  \caption{Qualitative amodal completion results on the HOD dataset. Our method can also accurately obtain amodal masks for objects with complex shapes}
  \figlabel{fig_amodal2}
\end{figure*}

\begin{figure*}[!ht]
    \centering
    \begin{subfigure}{\textwidth}
        \includegraphics[width=\textwidth,trim={1cm 0.6cm 0.2cm 0.6cm},clip]{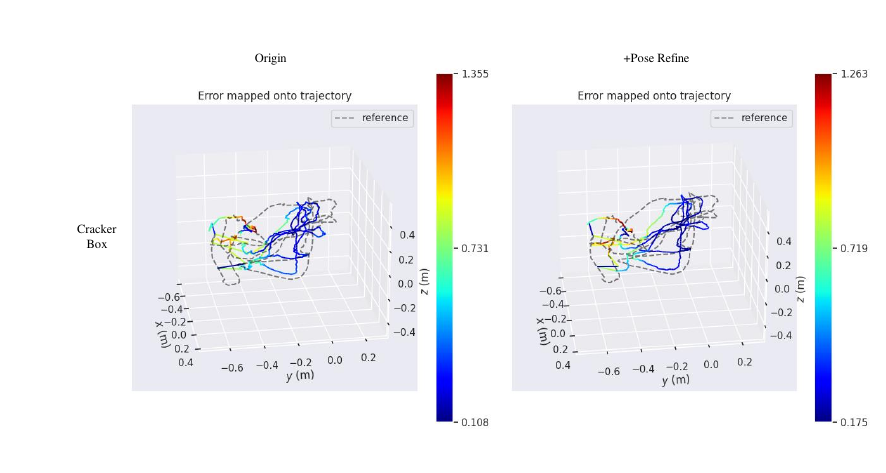}
    \end{subfigure}
\end{figure*}
\begin{figure*}[!ht]\ContinuedFloat
    \centering
    \begin{subfigure}{\textwidth}
        \includegraphics[width=\textwidth,trim={1.1cm 0.8cm 0.2cm 0.6cm},clip]{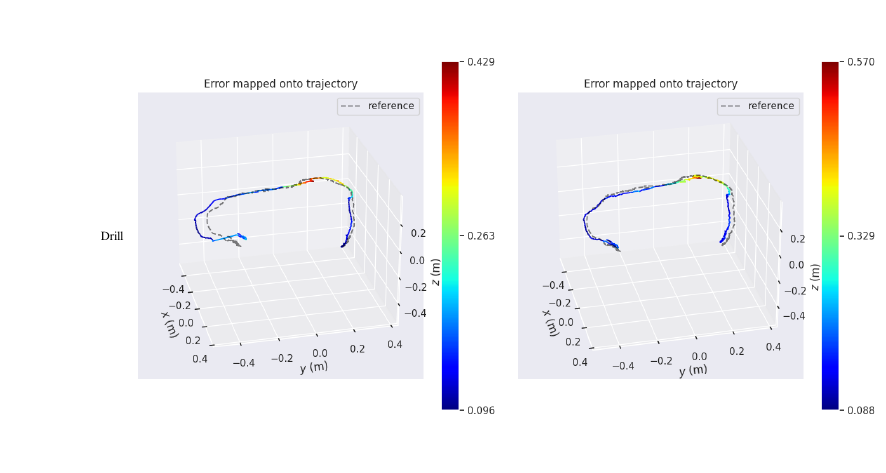}
    \end{subfigure}
\end{figure*}
\begin{figure*}[!ht]\ContinuedFloat
    \centering
    \begin{subfigure}{\textwidth}
        \includegraphics[width=\textwidth,trim={1.1cm 0.8cm 0.2cm 0.6cm},clip]{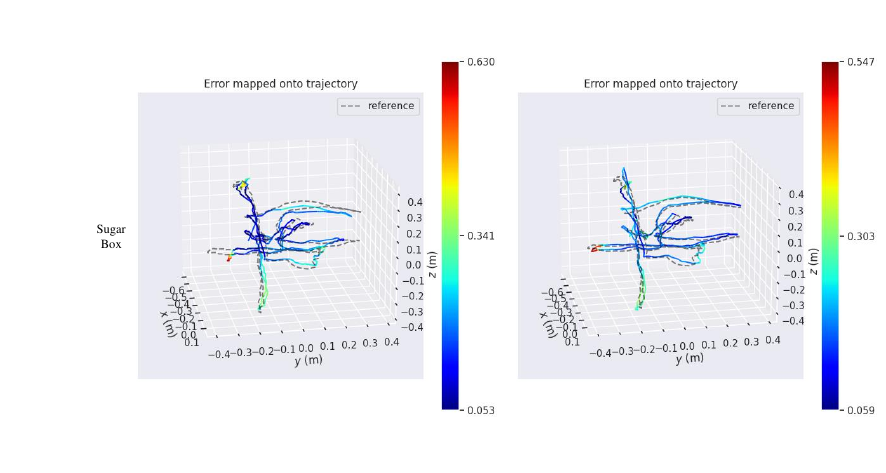}
    \end{subfigure}
\end{figure*}
\begin{figure*}[!ht]\ContinuedFloat
    \centering
    \begin{subfigure}{\textwidth}
        \includegraphics[width=\textwidth,trim={1.1cm 0.8cm 0.2cm 0.6cm},clip]{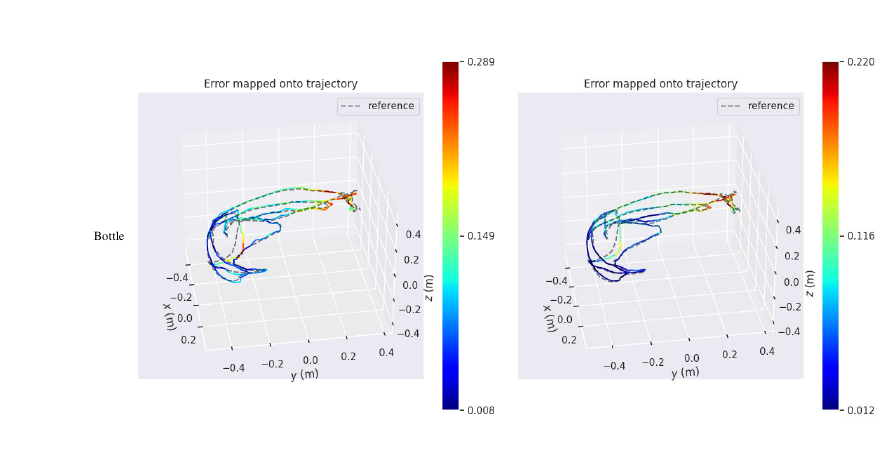}
    \end{subfigure}
\end{figure*}
\begin{figure*}[!ht]\ContinuedFloat
    \centering
    \begin{subfigure}{\textwidth}
        \includegraphics[width=\textwidth,trim={1.1cm 0.8cm 0.2cm 0.6cm},clip]{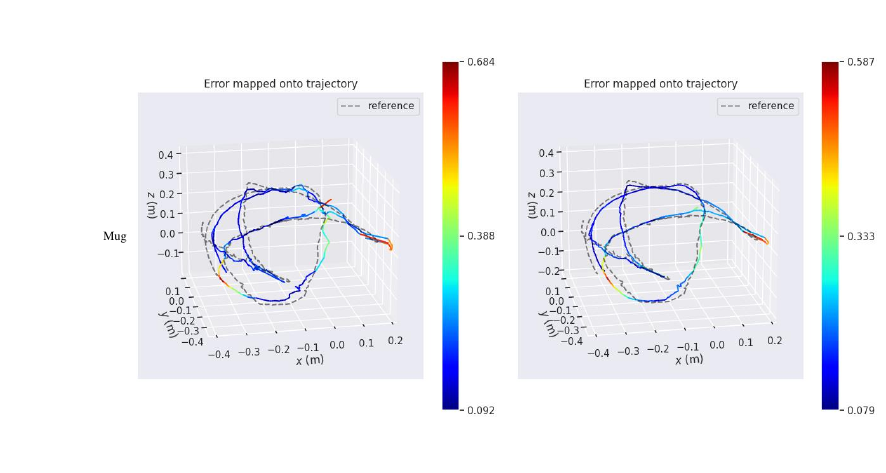}
    \end{subfigure}
    \caption{Error mapped onto trajectory on the HO3D dataset. Left: Raw results. Right: Results with pose refinement. Adding pose refinement can improve pose accuracy.}
    \figlabel{fig_motion}
\end{figure*}

\begin{figure*}[!ht]
    \centering
    \begin{subfigure}{\textwidth}
        \includegraphics[width=\textwidth,trim={1.3cm 1cm 0.2cm 0.6cm},clip]{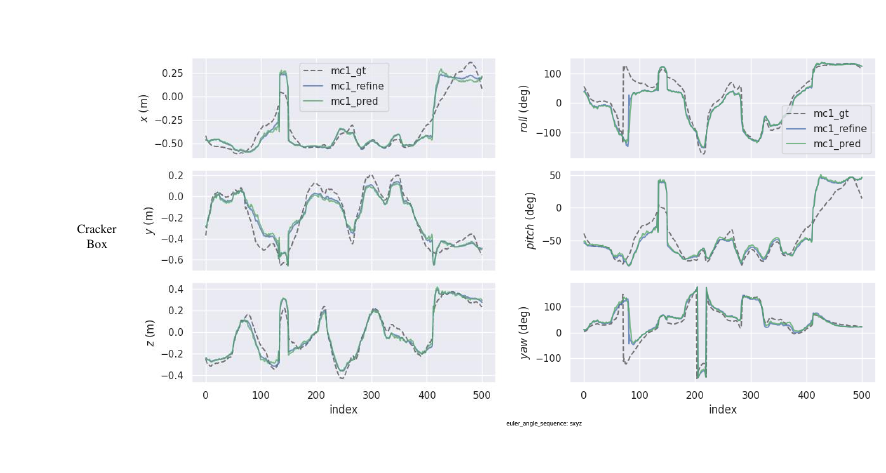}
    \end{subfigure}
\end{figure*}
\begin{figure*}[!ht]\ContinuedFloat
    \centering
    \begin{subfigure}{\textwidth}
        \includegraphics[width=\textwidth,trim={1.3cm 1cm 0.2cm 0.6cm},clip]{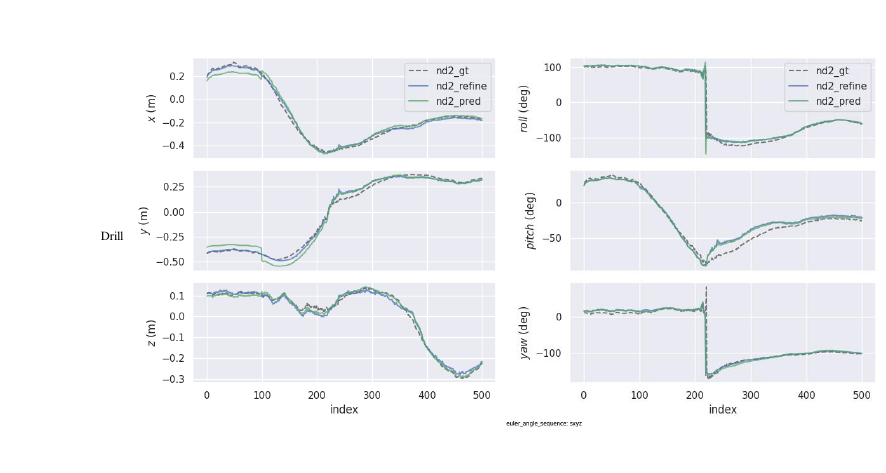}
    \end{subfigure}
\end{figure*}
\begin{figure*}[!ht]\ContinuedFloat
    \centering
    \begin{subfigure}{\textwidth}
        \includegraphics[width=\textwidth,trim={1.3cm 1cm 0.2cm 0.6cm},clip]{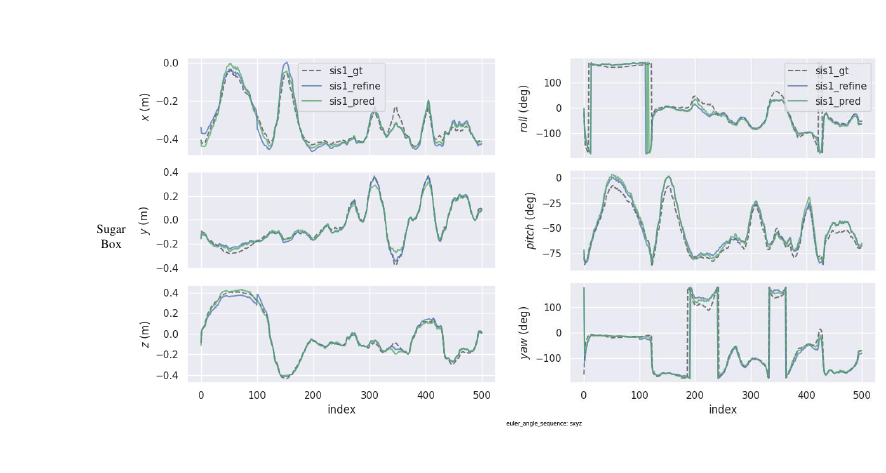}
    \end{subfigure}
\end{figure*}
\begin{figure*}[!ht]\ContinuedFloat
    \centering
    \begin{subfigure}{\textwidth}
        \includegraphics[width=\textwidth,trim={1.3cm 1cm 0.2cm 0.6cm},clip]{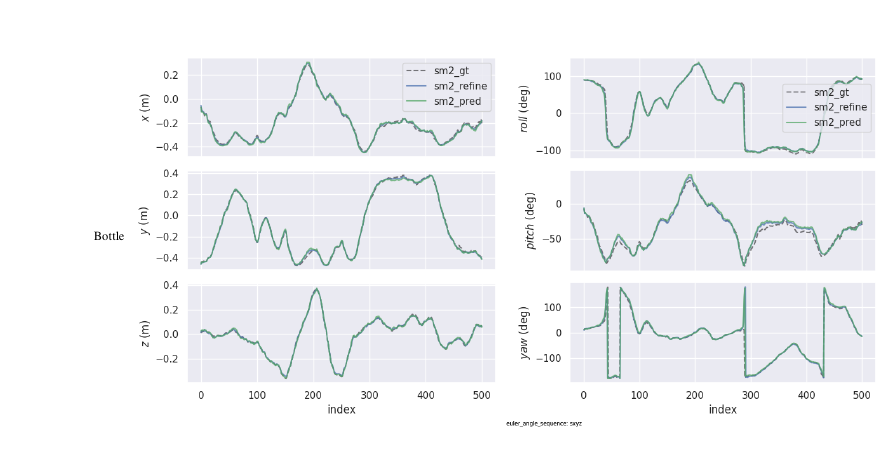}
    \end{subfigure}
\end{figure*}
\begin{figure*}[!ht]\ContinuedFloat
    \centering
    \begin{subfigure}{\textwidth}
        \includegraphics[width=\textwidth,trim={1.3cm 1cm 0.2cm 0.6cm},clip]{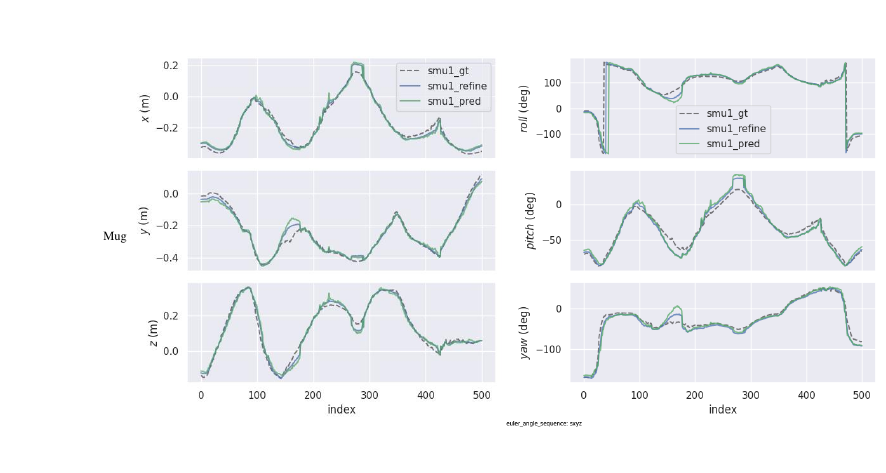}
    \end{subfigure}
    \caption{Trajectory against frame index on the HO3D dataset. Left: Translation. Right: Rotation.}
    \figlabel{fig_motion2}
\end{figure*}
\end{appendices}

\end{document}